\documentclass[letterpaper, 10 pt, conference]{ieeeconf}  

\IEEEoverridecommandlockouts                              

\usepackage{times} 
\usepackage{amsmath} 
\usepackage{amssymb}  
\usepackage{tabularray}
\usepackage{graphicx}
\usepackage{array}
\usepackage{multirow}
\usepackage{xcolor}

\DeclareMathOperator*{\argmax}{argmax}

\title{\LARGE \bf
Humanoids for Robot-Assisted Surgery: Bimanual Base Placement and Tool-Mount Optimization via Capability Maps
}

\author{Peihan Zhang$^{1}$, Zekai Liang$^{1}$, Florian Richter$^{1}$, Nikita Thareja$^{2}$, Ryan Broderick$^{2}$, Shanglei Liu$^{2}$,
\\ Michael Yip$^{1}$, \IEEEmembership{Senior Member, IEEE}
\thanks{$^{1}$Jacobs School of Engineering, UC San Diego, 9500 Gilman Drive, La Jolla, CA 92093, USA. {\tt\footnotesize\{pez004, z9liang, frichter, m1yip\} @ucsd.edu}}%
\thanks{$^2$Department of Surgery, UC San Diego, 9300 Campus Point Drive, La Jolla, CA 92037, USA. {\tt\footnotesize\{nthareja, rbroderick, s5liu\} @health.ucsd.edu}}%
}

\begin{document}

\maketitle
\thispagestyle{empty}
\pagestyle{empty}

\begin{abstract}

Rapid advances in humanoid robotics have motivated growing interest in the application of humanoids for healthcare and clinical tasks. However, it remains unclear how close contemporary humanoids are to meeting the kinematic demands of robot-assisted laparoscopic surgery. 
In this work, we address the question of optimal robot positioning through a quantitative analysis of workspace and robot setup configurations. We present a capability-map-based robot setup framework that optimizes humanoid base placement and tool mounting orientation to maximize bimanual humanoid reachability while accounting for tool-tip kinematics and remote-center-of-motion (RCM) constraints. We evaluate three humanoid platforms spanning different body dimensions and kinematic redundancy on workspace reachability for three representative general surgery procedures: cholecystectomy, inguinal hernia repair, and sleeve gastrectomy. The proposed joint optimization of base placement and tool mounting consistently outperforms base-only optimization and heuristic baselines. For cholecystectomy and inguinal hernia repair, which are characterized by relatively small and minimally overlapping workspaces, humanoid reachability approached $\sim$90\%. For the larger, overlapping multi-port arm workspace of sleeve gastrectomy, humanoids yield substantially lower coverage. These results quantify the near-term promise of humanoids for selected laparoscopic procedures and clarify key limitations that must be addressed for broader deployment.

\end{abstract}

\section{INTRODUCTION}

Robotic systems have become increasingly prevalent in medical and procedural applications, supporting tasks ranging from image-guided intervention to bedside logistics and rehabilitation~\cite{liangAutonomous2025}. Among these, {Robot-Assisted Surgery} (RAS) has emerged as a particularly impactful clinical application. RAS enables enhanced dexterity and increased motion precision during minimally invasive procedures, contributing to improved patient outcomes.
Recent rapid progress in embodied humanoid robotics and whole-body manipulation has motivated exploratory studies on applying humanoid systems to medical tasks.
While reports of humanoids in broader healthcare contexts are increasingly common~\cite{silvera-tawilRobotics2024}, it remains unclear how close contemporary general-purpose humanoids are to meeting RAS requirements.
\begin{figure}
    \centering
    \includegraphics[width=0.95\linewidth]{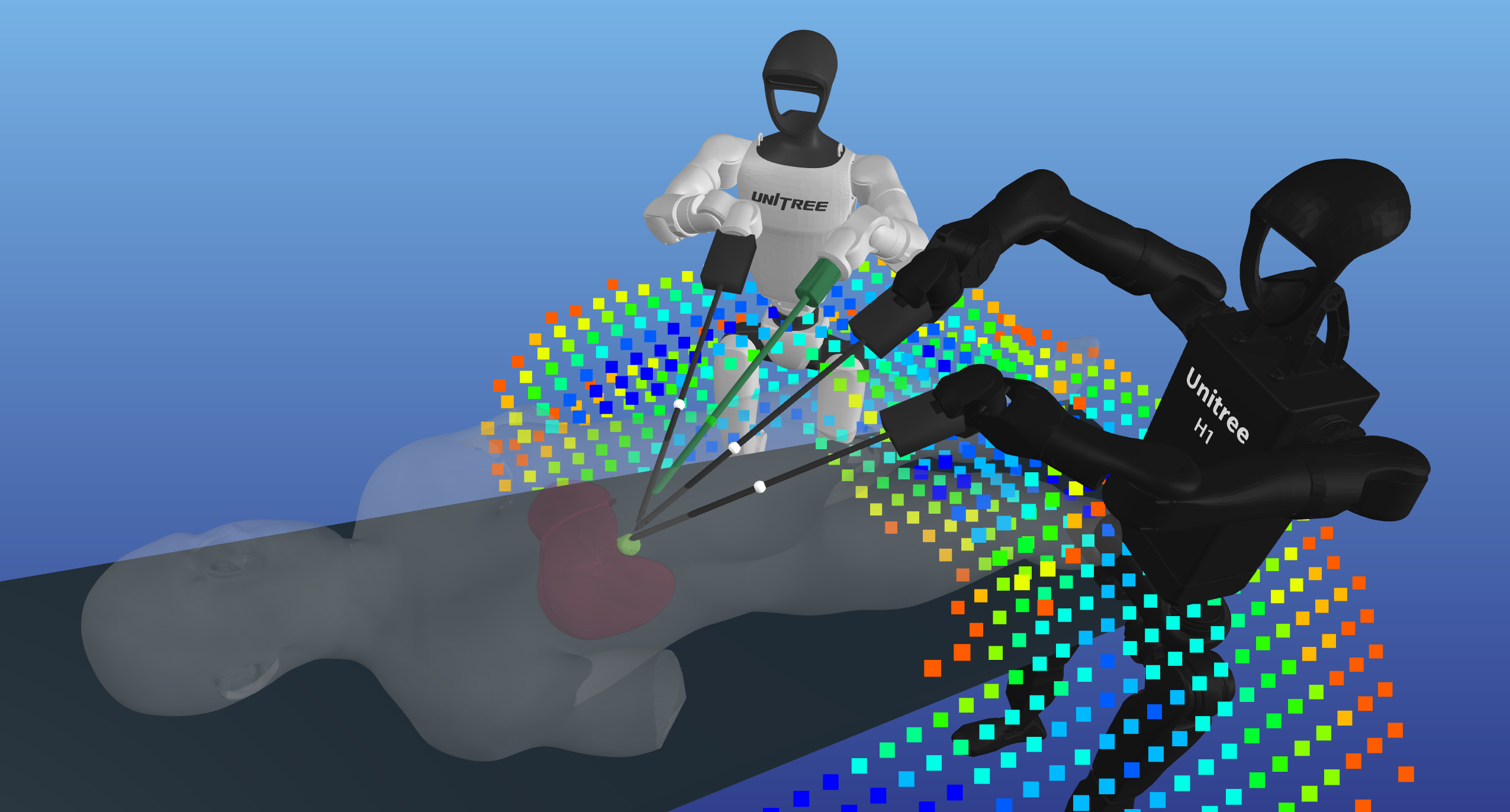}
    \caption{Cholecystectomy operated by two humanoids Unitree G1 and H1-2 with base placement for optimal port reachability. Floor tiles indicate base candidates colored by reachable-target count (blue high, red low).}
    \label{fig:cover-fig}
\end{figure}
For RAS, {preoperative setup and planning} are critical determinants of intraoperative feasibility and outcomes.
In particular, the selection of appropriate robot base placement, port locations and instrument approach directions directly shapes tool-tip workspace, dexterity, and collision-free accessibility throughout a procedure~\cite{hanARport2026}.
In current clinical practice, these decisions are largely guided by surgeon experience and iterative adjustments. As a result, the resulting port placement and instrument configurations tend to be platform-specific and difficult to generalize across systems or procedures.
This highlights the need for systematic, model-based assessments of reachability and workspace for more objective comparison across platforms for a given procedure.

Capability maps and their inversions have enabled efficient reachability querying and base placement for general-purpose manipulators in task planning and grasping~\cite{zachariasCapturing2007, vahrenkampRobot2013, zachariasCapability2013}. These have even been applied to early humanoid robot stance selection under simplified self-collision constraints~\cite{burgetStance2015}. However, these formulations do not transfer directly to RAS, where feasibility is dominated by operating-room obstacles and procedure-specific access constraints~\cite{sundaramTaskspecific2022}. More importantly, surgical effectiveness is determined by tool-tip reachability rather than the robot end-effector alone. The attached instrument’s geometry and articulation can significantly reshape the reachable workspace, and the transformation between the robot end-effector and tool base can vary depending on the mounting configuration~\cite{zhangConfiguration2021, osburgIncreasing2025, trabelsiRobot2024}. Recomputing an end-to-end capability map for every tool and tip frame is therefore impractical, motivating modular formulations that explicitly decouple robot and tool kinematics while respecting clinical constraints.

Rapid progress in humanoid robotics has renewed interest in the application of general-purpose anthropomorphic systems to clinical tasks.
Unlike dedicated surgical robots, humanoids are reconfigurable and low-footprint, can be repurposed across tasks and potentially improve access in resource-limited or space-constrained settings.
Although humanoids have been explored for hospital logistics, patient interaction, and educational roles, their suitability for surgical manipulation remains largely unquantified. The Surgie study~\cite{atarHumanoids2025} provides an early evaluation of humanoids performing dexterous clinical tasks under controlled conditions and highlights potential advantages for operating-room deployment. More recent work demonstrates the humanoid's capability of teleoperated laparoscopic manipulations~\cite{liangLapSurgie2026}, although no actual surgical procedures were performed. 

These early studies demonstrate the feasibility of teleoperated precision tasks and laparoscopic manipulation under controlled conditions, but a systematic evaluation of whether contemporary humanoids can satisfy the geometric and kinematic demands of representative surgical procedures is still lacking. A central question therefore remains: Can current humanoid robots achieve the reachability required for common laparoscopic surgeries, and under what conditions?
Addressing this question requires systematic evaluation of representative clinical procedures, supported by setup methods that explicitly account for the coordinated bimanual access and reachability required of humanoid platforms.

We propose a capability-map-based optimization framework for humanoid RAS setup planning that accounts for bimanual coordination and attached tool kinematics under procedure-specific constraints. 
Specifically, the framework:
(i) obtains humanoid capability map efficiently by exploiting its arm symmetry;
(ii) maps desired tool-tip targets to RCM-compliant tool-base poses via explicit tool kinematics; and
(iii) jointly optimizes humanoid base placement and tool mounting orientation to maximize aggregated bimanual tool-tip reachability over the task region.

The contributions of this paper are:
(i) a modular, capability-map-based formulation that decouples tool kinematics/RCM constraints from robot reachability and jointly optimizes humanoid base placement and tool mounting orientation to maximize bimanual reachability;
(ii) a systematic evaluation on three humanoid platforms spanning a wide range of body dimensions and kinematic redundancy across three representative robotic laparoscopic procedures with varying workspace complexity; and
(iii) an empirical assessment of procedures that appear most suitable for humanoid deployment, along with practical limitations that must be addressed for more complex multi-port surgery.



\section{RELATED WORK}
\label{sec:related_work}
Based on robot platforms and application scenarios, we group related work into three categories: general-purpose manipulators, medical robots, and humanoid systems.

\textbf{Base placement for general-purpose manipulators.}
Capability maps were originally introduced as discretized task-space representations to characterize a robot's ability to reach specific end-effector poses~\cite{zachariasCapturing2007}.
Building on these representations, inverse capability maps enabled efficient base-placement queries given a desired set of end-effector targets~\cite{vahrenkampRobot2013}.
Several works further enhanced reachability analysis beyond the end-effector by explicitly modeling end-effector-tool frame relationships. In particular, orientation-based reachability maps supported online end-effector frame extensions, allowing adaptation to different tool frames without requiring recomputation of the entire map~\cite{dongOrientationbased2015}.
More recently, learning-based approaches incorporate reachability information as prior knowledge to guide policy learning in mobile manipulation~\cite{jauhriRobot2022}.
While these methods provide valuable tools for general-purpose manipulators, they do not transfer directly to robot-assisted surgery (RAS). In surgical settings, feasibility is governed not only by end-effector reachability but also by procedure- and environment-specific constraints (e.g., RCM constraints and patient anatomy) as well as by the geometry and additional degrees of freedom of surgical instruments~\cite{sundaramTaskspecific2022}.

\textbf{Robot Setup in Medical Tasks.}
Robot setup in medical scenarios is typically framed as an optimization problem in which reachability metrics are maximized under task- and environment-specific constraints.
Within multi-port surgical systems, \cite{xuOptimal2021} proposes a multi-criteria optimization that balances reachability with self- and environment-collision avoidance. 
\cite{trabelsiRobot2024} optimizes accessibility and manipulability under explicit RCM constraints for an endoscope holder system.
To account for procedure-dependent constraints, \cite{sundaramTaskspecific2022} has adapted capability-map to surgical contexts by incorporating laparoscopic RCM points and kinematic restrictions. Preoperative base/access-point optimization is achieved via multi-objective genetic algorithms across multiple surgical use cases. However, \cite{sundaramTaskspecific2022} iterates over a large set of base and RCM configurations, which can be computationally expensive and produces maps tied to specific robot–tool configurations that limits its use for different instruments.
Beyond purely model-based optimization, \cite{yoonOptimizing2024} used data-driven clustering of recorded surgeon motions to infer preferred pose distributions of robot base placement.
Despite these advances, many approaches treat the tool as rigidly attached to the end-effector. However, in practice, the end-effector-to-tool transformation can vary across tools and can substantially impact task-space reachability. For example, \cite{zhangConfiguration2021} demonstrates how two different endoscope installation choices affect reachable dexterity.
\cite{osburgIncreasing2025} demonstrated that allowing controlled variation in the end-effector--tool orientation significantly increases the set of high-reachability base poses for ultrasound scanning. In laparoscopic systems, \cite{trabelsiRobot2024} proposed multi-criteria optimization to simultaneously optimize robot base placement and laparoscope mounting orientation under RCM constraints .

\textbf{Humanoids for Surgery.}
Recent advances in robot learning and humanoid hardware have enabled increasingly human-like dexterity and autonomy, raising the prospect of embodied humanoids performing sophisticated tasks that have traditionally required human operators.
In healthcare, most humanoid-oriented efforts have focused on non-surgical applications such as logistics, patient interaction, and education~\cite{kyrariniSurvey2021, silvera-tawilRobotics2024}.
In contrast, research on humanoids in surgical contexts remains limited.
~\cite{atarHumanoids2025} investigates the feasibility of teleoperated humanoids for direct clinical tasks, including precision needle tasks, using the Unitree G1 platform.
More recently, ~\cite{liangLapSurgie2026} proposes a humanoid-based laparoscopic teleoperation framework to leverage off-the-shelf laparoscopic instruments.
However, these studies do not address the base-placement problem for humanoids, and a systematic evaluation of humanoid capability for representative surgical procedures remains largely unexplored.


\section{METHOD}

\subsection{Problem Formulation}
\label{sec:problem_formulation}

\begin{figure}
    \centering
    \includegraphics[width=0.9\linewidth]{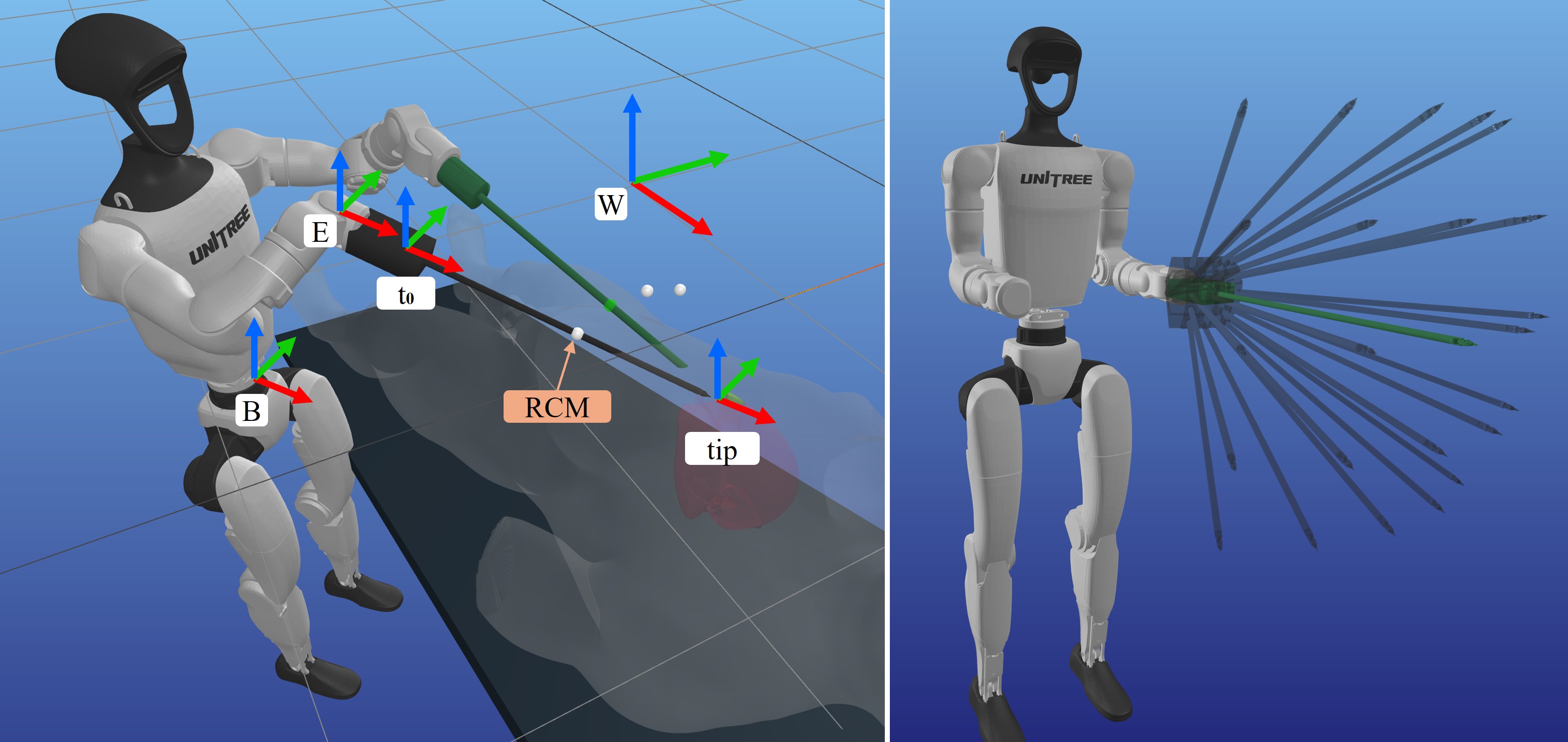}
    \caption{Coordinate frames (left) and tool mount scheme (right) illustration.}
    \label{fig:frame-mounting}
\end{figure}

Consider humanoid robots operating the surgery at a given workspace. Surgical tools are attached to two end-effectors of the robot arms.
We denote the coordinate frames as $\mathrm{W}$ (world), $\mathrm{B}$ (robot base), $\mathrm{E}$ (end-effector), $\mathrm{t}_0$ (tool base), and $\mathrm{tip}$ (tool tip) as shown in Fig.~\ref{fig:frame-mounting}.
Let $SO(3)$ be the special orthogonal group, 
$SE(3)$ the special Euclidean group, $S^1\triangleq \{\psi\in \left[0, 2\pi \right) \}$ the unit circle.
${}^{\mathrm{X}}\!\mathbf{T}_{\mathrm{Y}}\in SE(3)$ denotes the homogeneous transformation from frame $\mathrm{Y}$ to $\mathrm{X}$.
The tool tip pose can be express by kinematic chain
\begin{equation}
{}^{\mathrm{W}}\!\mathbf{T}_{\mathrm{tip}}
=
{}^{\mathrm{W}}\!\mathbf{T}_{\mathrm{B}}(\mathbf{b})
{}^{\mathrm{B}}\mathbf{T}_{\mathrm{E}}(\mathbf{q}_r)
{}^{\mathrm{E}}\mathbf{T}_{\mathrm{t}_0}(\boldsymbol{\eta})
{}^{\mathrm{t}_0}\mathbf{T}_{\mathrm{tip}}(\mathbf{q}_t).
\label{eq:chain}
\end{equation}
where the robot base pose is defined as
\begin{equation}
\mathbf{b} \triangleq (x,y,z,\psi)\in \mathcal{S}_{\mathrm{B}}\subset \mathbb{R}^3\times S^1
\label{eq:set_base}
\end{equation}
within admissible surgical region $\mathcal{S}_{\mathrm{B}}$ with associated transform ${}^{\mathrm{W}}\!\mathbf{T}_{\mathrm{B}}(\mathbf{b})\in SE(3)$. 
$\mathbf{q}_r\in\mathcal{C}_r$ and $\mathbf{q}_t\in\mathcal{C}_t$ denote robot arm joints and tool joints configurations, and robot arm and tool forward kinematics are written as ${}^{\mathrm{B}}\mathbf{T}_{\mathrm{E}}(\mathbf{q}_r)$ and ${}^{\mathrm{t}_0}\mathbf{T}_{\mathrm{tip}}(\mathbf{q}_t)$, respectively. 
Consider the tool mounted on the end-effector has adjustable orientations as shown in Fig.~\ref{fig:frame-mounting} parameterized by $\boldsymbol{\eta}$ with associated transform
\begin{equation}
{}^{\mathrm{E}}\mathbf{T}_{\mathrm{t}_0}(\boldsymbol{\eta})\in \mathcal{S}_{\mathrm{ori}} \subset SE(3),
\end{equation}

A robot capability map which encodes the reachability of end-effector poses is a binary occupancy function
\begin{equation}
\begin{aligned}
&\mathcal{M}\left(\cdot\right):\ SE(3) \rightarrow\{0,1\}, \\
& N_r \triangleq \left|\left\{k\in \mathcal{S}_{{\mathrm{E}}}\mid \mathcal{M}(k)=1\right\}\right|,
\end{aligned}
\end{equation}
where $\mathcal{M}\left({}^{\mathrm{B}}\mathbf{T}_{\mathrm{E}}\right)=1$ indicates a pose ${}^{\mathrm{B}}\mathbf{T}_{\mathrm{E}} $ in interested end-effector working region  $\mathcal{S}_{{\mathrm{E}}}$ is reachable with some $\mathbf{q}_r$.


Let $N_t$ desired tool-tip pose targets be
$\mathcal{S}_{\mathrm{tip}}=\{{}^{\mathrm{W}}\!\mathbf{T}^{\star}_{\mathrm{tip},i}\}_{i=1}^{N_t}$.
Using the notation ${}^{X}\mathbf{T}_{Y}^{-1} \triangleq ({}^{X}\mathbf{T}_{Y})^{-1}$, 
inverting~\eqref{eq:chain} yields robot-base pose candidates
\begin{equation}
\begin{aligned}
{}^{\mathrm{W}}\!\mathbf{T}_{\mathrm{B}, i, j}
& = 
{}^{\mathrm{W}}\!\mathbf{T}^{\star}_{\mathrm{tip},i}
{}^{\mathrm{t}_0}\mathbf{T}_{\mathrm{tip}}^{-1}
{}^{\mathrm{E}}\mathbf{T}_{\mathrm{t}_0}^{-1}(\boldsymbol{\eta})
{}^{\mathrm{B}}\mathbf{T}_{\mathrm{E}, j}^{-1} \\
\text{s.t.}\quad
& i \in \{1,\dots,N_t\}, \quad j \in \{1,\dots,N_r\}, \\
& \mathcal{M}\left({}^{\mathrm{B}}\mathbf{T}_{\mathrm{E},j}\right)=1, \\
& g_{\mathrm{rcm}}\!\left({}^{\mathrm{W}}\!\mathbf{T}^{\star}_{\mathrm{tip},i}
{}^{\mathrm{t}_0}\mathbf{T}_{\mathrm{tip}}^{-1}, \,{}^{\mathrm{W}}\!\mathbf{p}_{\mathrm{rcm}}\right)=0.
\end{aligned}
\label{eq:chain-inverse}
\end{equation}

where ${}^{\mathrm{W}}\!\mathbf{p}_{\mathrm{rcm}}\in\mathbb{R}^3$ denotes the RCM point, and $g_{\mathrm{rcm}}(\cdot)$ indicates whether the tool satisfies the RCM constraint.

We seek $(\mathbf{b},\boldsymbol{\eta})$ that maximizes the aggregated reachability of these targets at robot base pose $\mathbf{b}$ and tool mounting $\boldsymbol{\eta}$:
\begin{equation}
\begin{aligned}
(\mathbf{b}^{\star},\boldsymbol{\eta}^{\star})
& =
\argmax_{\mathbf{b}\in\mathcal{S}_{\mathrm{B}},~\boldsymbol{\eta}\in\mathcal{S}_{\mathrm{ori}}}
\ \sum_{i=1}^{N_t} \sum_{j=1}^{N_r}
\Phi\!\left({}^{\mathrm{W}}\!\mathbf{T}_{\mathrm{B}, i, j}\right),
\label{eq:opt}
\end{aligned}
\end{equation}
where $\Phi(\cdot)$ is a reachability score function.

\subsection{Robot Capability Map and its Inversion}
\label{sec:cmap_generation}
The robot capability map is constructed offline following procedures in \cite{porgesReachability2014}.
Specifically, we voxelize the arm workspace in discretized $SE(3)$ as 5D grid (3D positions, approaching directions, and rolls), sample joint configurations $\mathbf{q}_r\in\mathcal{Q}_r$, evaluate forward kinematics ${}^{\mathrm{B}}\mathbf{T}_{\mathrm{E}}(\mathbf{q}_r)$ from the robot base to the end-effector, and label the corresponding grid of feasible collision-free poses.
We define a discretization operator $\mathcal{D}(\cdot)$ that maps a pose in $SE(3)$ to a unique bin index,
\begin{equation}
k=\mathcal{D}\!\left({}^{\mathrm{B}}\mathbf{T}_{\mathrm{E}}\right)\in\mathcal{K}, 
\quad
\mathcal{K}\triangleq\{1,\dots, N_{bin} \},
\end{equation}
where $N_{bin} $ is the total number of map voxel bins.



Without loss of generality, we assume the humanoid has two identical arms symmetric with respect to the $x$--$z$ plane of the base frame.
Let $\mathcal{M}^{\mathrm{L}}$ denote the left-arm capability map, represented as a binary occupancy function
\begin{equation}
\mathcal{M}^{\mathrm{L}}:\mathcal{K}\rightarrow\{0,1\},
\end{equation}
where $\mathcal{M}^{\mathrm{L}}(k)=1$ indicates that there exists a collision-free left-arm configuration
$\mathbf{q}_r^{\mathrm{L}}\in\mathcal{Q}_r^{\mathrm{L}}$ such that the end-effector pose ${}^{\mathrm{B}}\mathbf{T}_{\mathrm{E}}(\mathbf{q}_r^{\mathrm{L}})$
falls into bin $k$ (i.e., $k=\mathcal{D}({}^{\mathrm{B}}\mathbf{T}_{\mathrm{E}}(\mathbf{q}_r^{\mathrm{L}}))$).

In \eqref{eq:chain-inverse}, the inverse of all reachable entries in the capability map is used, which can be precomputed offline and stored as an inverse map for efficient base-placement queries in practice.
Given the reachable left-arm end-effector poses
\begin{equation}
\mathcal{S}_{\mathrm{map}}^{\mathrm{L}} \triangleq \left\{\,{}^{\mathrm{B}}\mathbf{T}_{\mathrm{E}}\in SE(3)\ \middle|\ \mathcal{M}^{\mathrm{L}}(\mathcal{D}({}^{\mathrm{B}}\mathbf{T}_{\mathrm{E}}))=1\,\right\},
\end{equation}
its inverse map is
\begin{equation}
\mathcal{S}_{{\mathrm{map}}^{-1}}^{\mathrm{L}} \triangleq \left\{\,\left({}^{\mathrm{B}}\mathbf{T}_{\mathrm{E}}\right)^{-1}\ \middle|\ {}^{\mathrm{B}}\mathbf{T}_{\mathrm{E}}\in \mathcal{S}_{\mathrm{map}}^{\mathrm{L}}\,\right\}.
\label{eq:inv_map_set}
\end{equation}

Using arm symmetry assumption, we query right-arm reachability from $\mathcal{M}^{\mathrm{L}}$ without constructing $\mathcal{M}^{\mathrm{R}}$.
Let
\begin{equation}
\mathbf{S}\triangleq
\begin{bmatrix}
\mathrm{diag}(1,-1,1) & \mathbf{0}\\
\mathbf{0}^\top & 1
\end{bmatrix}
\end{equation}
denote the reflection about the $x$--$z$ plane (i.e., $y\mapsto -y$). Then, the right-arm map and inverse map satisfies
\begin{equation}
\mathcal{M}^{\mathrm{R}}\!\left(\mathcal{D}\!\left({}^{\mathrm{B}}\mathbf{T}_{\mathrm{E}}\right)\right)
=
\mathcal{M}^{\mathrm{L}}\!\left(\mathcal{D}\!\left(\mathbf{S}\,{}^{\mathrm{B}}\mathbf{T}_{\mathrm{E}}\,\mathbf{S}^{-1}\right)\right),
\label{eq:right_from_left_symmetry}
\end{equation}
\begin{equation}
\mathcal{S}_{{\mathrm{map}}^{-1}}^{\mathrm{R}} = \left\{\,\mathbf{S}\,{}^{\mathrm{B}}\mathbf{T}^{-1}_{\mathrm{E}}\,\mathbf{S^{-1}}\ \middle|\ {}^{\mathrm{B}}\mathbf{T}^{-1}_{\mathrm{E}}\in \mathcal{S}_{{\mathrm{map}}^{-1}}^{\mathrm{L}}\,\right\}.
\label{eq:right_inv_from_left}
\end{equation}

Since surgical feasibility depends on tool-tip rather than end-effector poses, an end-to-end robot--tool capability map is possible but must be recomputed per tool. We instead model tool kinematics and RCM constraints separately in the next subsection, decoupling the robot map from the tool and enabling reuse across tools.

\subsection{Inversion of Tool-tip Targets with RCM constraints}
\label{sec:task_backprop_tool}
The surgical task space is specified as a finite set of desired tool-tip poses $\mathcal{S}_{\mathrm{tip}}$, whereas the robot capability map is defined over end-effector poses at the tool base.
To bridge this mismatch, we map each tool-tip target to a candidate tool-base pose which satisfies task-specific RCM constraints.

In robot-assisted surgery, the RCM requires the instrument shaft to pass through the trocar point; we assume the tool-base $x$-axis is aligned with the shaft.
Let ${}^{\mathrm{W}}\!\mathbf{T}_{\mathrm{t}_0}=\begin{bmatrix}{}^{\mathrm{W}}\!\mathbf{R}_{\mathrm{t}_0} & {}^{\mathrm{W}}\!\mathbf{p}_{\mathrm{t}_0}\\ \mathbf{0}^\top & 1\end{bmatrix}$.
The unit shaft direction is then
\begin{equation}
{}^{\mathrm{W}}\!\mathbf{a}
\triangleq
{}^{\mathrm{W}}\!\mathbf{R}_{\mathrm{t}_0}\mathbf{e}_x,
\qquad \|{}^{\mathrm{W}}\!\mathbf{a}\|=1,
\label{eq:shaft_axis_from_T0}
\end{equation}
where $\mathbf{e}_x=[1~0~0]^\top$.
We model the RCM compliance by the collinearity of the shaft line through ${}^{\mathrm{W}}\!\mathbf{p}_{\mathrm{t}_0}$ and ${}^{\mathrm{W}}\!\mathbf{p}_{\mathrm{rcm}}$:
\begin{equation}
g_{\mathrm{rcm}}\!\left({}^{\mathrm{W}}\!\mathbf{T}_{\mathrm{t}_0},{}^{\mathrm{W}}\!\mathbf{p}_{\mathrm{rcm}}\right)
\triangleq
\left({}^{\mathrm{W}}\!\mathbf{p}_{\mathrm{rcm}}-{}^{\mathrm{W}}\!\mathbf{p}_{\mathrm{t}_0}\right)\times {}^{\mathrm{W}}\!\mathbf{a}
=
\mathbf{0},
\label{eq:g_rcm_T0}
\end{equation}
equivalently, there exists an insertion scalar $s\in\mathbb{R}$ such that
\begin{equation}
{}^{\mathrm{W}}\!\mathbf{p}_{\mathrm{rcm}}
=
{}^{\mathrm{W}}\!\mathbf{p}_{\mathrm{t}_0} + s\,{}^{\mathrm{W}}\!\mathbf{a},
\quad s\in[s_{\min},s_{\max}],
\label{eq:rcm_insertion_T0}
\end{equation}
where $s$ corresponds to insertion depth along the shaft axis and $[s_{\min},s_{\max}]$ is bounded by instrument geometry and clinical limits.
In practice, we allow a small tolerance $\epsilon_{\mathrm{rcm}}$ by enforcing $\|g_{\mathrm{rcm}}(\cdot)\|\le\epsilon_{\mathrm{rcm}}$.

Another constraint is given by tool kinematics: 
\begin{equation}
\exists\,\mathbf{q}_t\in\mathcal{Q}_t, \quad
{}^{\mathrm{W}}\!\mathbf{T}^{\star}_{\mathrm{tip}}
=
{}^{\mathrm{W}}\!\mathbf{T}_{\mathrm{t}_0}\;{}^{\mathrm{t}_0}\mathbf{T}_{\mathrm{tip}}(\mathbf{q}_t).
\label{eq:tip_consistency}
\end{equation}

Based on tool kinematics, we distinguish between passive and active instruments. Passive tools are rigid instruments such as endoscopes or ultrasound probes. 
Active tools are articulated instruments with internal joints:
\begin{equation}
{}^{\mathrm{t}_0}\mathbf{T}_{\mathrm{tip}}(\mathbf{q}_t)=
\begin{cases}
\bar{\mathbf{T}}_{\mathrm{t}_0,\mathrm{tip}}, & \text{passive tool},\\[2pt]
FK_t(\mathbf{q}_t),\ \mathbf{q}_t\in\mathcal{Q}_t, & \text{active tool}.
\end{cases}
\label{eq:tool_fk_piecewise}
\end{equation}


For passive tools, \eqref{eq:tip_consistency} gives a unique tool-base pose ${}^{\mathrm{W}}\!\mathbf{T}_{\mathrm{t}_0}$ for each ${}^{\mathrm{W}}\!\mathbf{T}^{\star}_{\mathrm{tip}}$, with the tip $x$-axis aligned with the shaft. Thus, rather than filtering invalid poses, we can sample ${}^{\mathrm{W}}\!\mathbf{T}^{\star}_{\mathrm{tip}}$ that already satisfy \eqref{eq:g_rcm_T0} and recover ${}^{\mathrm{W}}\!\mathbf{T}_{\mathrm{t}_0}$ via \eqref{eq:tip_consistency}.
For active tools, multiple $\mathbf{q}_t$ may realize the same tip target, and distal articulation implies that the tip frame need not align with the shaft axis; therefore, RCM must be enforced via the tool-base shaft axis as in \eqref{eq:g_rcm_T0}, while tip consistency is enforced via \eqref{eq:tip_consistency}. We enumerate or solve for feasible $\mathbf{q}_t$ and retain those candidates that satisfy both constraints.

Finally, applying the above procedure to $\mathcal{S}_{\mathrm{tip}}$ yields RCM-compliant tool-base targets set
\begin{equation}
\begin{aligned}
\mathcal{S}^{\star}_{\mathrm{t}_0}
\triangleq
\left\{ \right.
{}^{\mathrm{W}}\!\mathbf{T}_{\mathrm{t}_0}^{\star}
\left.\middle| \right.
&
\exists\, i\in\{1,\dots,N_t\}, \exists\,\mathbf{q}_t\in\mathcal{Q}_t,\\
& {}^{\mathrm{W}}\!\mathbf{T}^{\star}_{\mathrm{tip},i} = {}^{\mathrm{W}}\!\mathbf{T}_{\mathrm{t}_0}^{\star}\;{}^{\mathrm{t}_0}\mathbf{T}_{\mathrm{tip}}(\mathbf{q}_t),\\
& 
\ \|g_{\mathrm{rcm}}({}^{\mathrm{W}}\!\mathbf{T}_{\mathrm{t}_0}^{\star},{}^{\mathrm{W}}\!\mathbf{p}_{\mathrm{rcm}})\|\le \epsilon_{\mathrm{rcm}}
\left. \right\}.
\label{eq:G_T0_star}
\end{aligned}
\end{equation}

These tool-base poses are then matched to the robot capability map through the tool-mounting transform.

\subsection{Base Placement Optimization with Tool Mounting Variation }
\label{subsec:opt-framework}

While the tool is attached to the robot end-effector, the mount configuration also affects task-space reachability. We assume a fixed mount translation and optimize its orientation. We discretize feasible orientations in roll--pitch--yaw (RPY) and define the candidate mount set as
\begin{equation}
\mathcal{S}_{\mathrm{ori}}
\triangleq
\left\{
\begin{bmatrix}
{}^{\mathrm{E}}\mathbf{R}_{\mathrm{t}_0}(\boldsymbol{\eta}) & \mathbf{t}_{\mathrm{E},\mathrm{t}_0}^{\star}\\
\mathbf{0}^\top & 1
\end{bmatrix}
\ \middle|\
\boldsymbol{\eta}=(\phi,\theta,\psi)\in \Phi\times\Theta\times\Psi
\right\},
\label{eq:Sori}
\end{equation}
where ${}^{\mathrm{E}}\mathbf{R}_{\mathrm{t}_0}(\boldsymbol{\eta})=\mathbf{R}_z(\psi)\mathbf{R}_y(\theta)\mathbf{R}_x(\phi)$ and $\Phi,\Theta,\Psi$ are discretized angle sets for roll, pitch, and yaw, respectively. 
$\mathbf{t}_{\mathrm{E},\mathrm{t}_0}^{\star}$ is fixed translation.


First consider the left arm of the humanoid robot. Given the discrete tool mounting set $\mathcal{S}_{\mathrm{ori}}^{\mathrm{L}}$ in \eqref{eq:Sori}, the RCM-compliant tool-base target set $\mathcal{S}^{\star, L}_{\mathrm{t}_0}$ in \eqref{eq:G_T0_star}, and the inverse robot capability map $\mathcal{S}_{{\mathrm{map}}^{-1}}^{\mathrm{L}}$ in \eqref{eq:inv_map_set}, the base placement candidate in \eqref{eq:chain-inverse} of the left arm can be written as
\begin{equation}
\begin{aligned}
& {}^{\mathrm{W}}\!\mathbf{T}_{\mathrm{B}, i, j, k}^{\mathrm{L}}
 = 
{}^{\mathrm{W}}\!\mathbf{T}^{\star}_{\mathrm{t}_0, i}\,
\left({}^{\mathrm{E}}\mathbf{T}_{\mathrm{t}_0, k}\right)^{-1}\,
{}^{\mathrm{B}}\mathbf{T}_{\mathrm{E}, j}^{-1}, \\
\text{s.t.}\quad
& {}^{\mathrm{W}}\!\mathbf{T}^{\star}_{\mathrm{t}_0, i} \in \mathcal{S}^{\star, L}_{\mathrm{t}_0}, i\in\{1,\dots,N_t\}, \\
&{}^{\mathrm{E}}\mathbf{T}_{\mathrm{t}_0, k} \in \mathcal{S}_{\mathrm{ori}}^{\mathrm{L}}, k\in\{1,\dots,N_{\mathrm{ori}}\}, \\
&{}^{\mathrm{B}}\mathbf{T}_{\mathrm{E}, j}^{-1} \in \mathcal{S}_{{\mathrm{map}}^{-1}}^{\mathrm{L}}, j\in\{1,\dots, \mid \mathcal{S}_{{\mathrm{map}}^{-1}}^{\mathrm{L}} \mid\}.
\end{aligned}
\label{eq:base_ijk}
\end{equation}

Let $N_{\mathrm{B}} \triangleq |\mathcal{S}_{\mathrm{B}}|$ be the number of discretized base bins of admissible robot base pose, 
$N_{\mathrm{ori}} \triangleq |\mathcal{S}_{\mathrm{ori}}|$ the number of discretized tool mounting orientations.
Let $\mathcal{D}_{\mathrm{B}}(\cdot)$ map a robot base pose ${}^{\mathrm{W}}\!\mathbf{T}_{\mathrm{B}}$ to a base-bin index, and $\mathcal{D}_{\mathrm{ori}}(\cdot)$ map a tool mounting orientation ${}^{\mathrm{E}}\mathbf{T}_{\mathrm{t}_0}$ to an orientation-bin index
\begin{align}
    m &= \mathcal{D}_{\mathrm{B}}({}^{\mathrm{W}}\!\mathbf{T}_{\mathrm{B}})\in\{1,\dots,N_{\mathrm{B}}\}, \\
    n &= \mathcal{D}_{\mathrm{ori}}({}^{\mathrm{E}}\mathbf{T}_{\mathrm{t}_0})\in\{1,\dots,N_{\mathrm{ori}}\}.
\end{align}

Define reachability score function $\Phi(\cdot)$ as a one-hot matrix
\begin{equation}
\Phi\!\left({}^{\mathrm{W}}\!\mathbf{T}_{\mathrm{B},i,j,k},\,{}^{\mathrm{E}}\mathbf{T}_{\mathrm{t}_0,k}\right)
\triangleq
\mathbf{e}_{\mathcal{D}_{\mathrm{B}}({}^{\mathrm{W}}\!\mathbf{T}_{\mathrm{B},i,j,k})}\ \mathbf{e}_{\mathcal{D}_{\mathrm{ori}}({}^{\mathrm{E}}\mathbf{T}_{\mathrm{t}_0,k})}^\top
,
\label{eq:Phi_matrix}
\end{equation}
where $\mathbf{e}_{\mathcal{D}_{\mathrm{B}}(\cdot)}\in\mathbb{R}^{N_{\mathrm{B}}}$ and $\mathbf{e}_{\mathcal{D}_{\mathrm{ori}}(\cdot)}\in\mathbb{R}^{N_{\mathrm{ori}}}$ are standard basis vectors.
Summing over all samples of left arm in \eqref{eq:base_ijk} yields the single-arm score matrix
\begin{equation}
\mathbf{H}^{\mathrm{L}}
\triangleq
\sum_{i, j, k}
\Phi\!\left({}^{\mathrm{W}}\!\mathbf{T}_{\mathrm{B},i,j,k}^{\mathrm{L}} \,,{}^{\mathrm{E}}\mathbf{T}_{\mathrm{t}_0,k}\right)
\in \mathbb{N}^{N_{\mathrm{B}}\times N_{\mathrm{ori}}}.
\label{eq:H_single}
\end{equation}


Given score matrix for both arm $\mathbf{H}^{\mathrm{L}}$ and $\mathbf{H}^{\mathrm{R}}$, we solve
\begin{equation}
(m^\star,n_{\mathrm{L}}^\star,n_{\mathrm{R}}^\star)
=
\argmax_{m, n_{\mathrm{L}}, n_{\mathrm{R}}}
\left(\mathbf{H}^L_{m,n_{\mathrm{L}}} + \mathbf{H}^R_{m,n_{\mathrm{R}}}\right),
\label{eq:decode_bimanual}
\end{equation}
where $m\in\{1,\dots,N_{\mathrm{B}}\}$ and $n_{\mathrm{L}},n_{\mathrm{R}}\in\{1,\dots,N_{\mathrm{ori}}\}$.
The optimal robot base and tool mounts are then decoded as
\begin{equation}
\mathbf{b}^\star=\mathcal{D}_{\mathrm{B}}^{-1}(m^\star),\
\boldsymbol{\eta}_{\mathrm{L}}^\star=\mathcal{D}_{\mathrm{ori}}^{-1}(n_{\mathrm{L}}^\star),\
\boldsymbol{\eta}_{\mathrm{R}}^\star=\mathcal{D}_{\mathrm{ori}}^{-1}(n_{\mathrm{R}}^\star).
\label{eq:opt-result}
\end{equation}

\section{EXPERIMENTS}
{
\subsection{Port Placement and Patient Position in Common Robotic Laparoscopic Procedures}
\label{sec:port_placement_common}

To approximate patient anatomy in simulation, we use publicly available Human Reference Atlas (HRA)~\cite{bornerHuman2025} 3D models of the skin surface, liver, gallbladder, and pelvis, corresponding to a 38-year-old male (180.3\,cm, 199\,lb).

We obtained feedback from two residency- and fellowship-trained general surgeons on port positioning for common general surgery procedures. Since port-placement guidelines for humanoid platforms are not yet established, we adopted practices reported for other robotic surgical systems (e.g., \textit{da Vinci}). We limited our analysis to single-quadrant procedures, as port placement and working space are considerably more variable in multi-quadrant cases. Because clinical configurations depend heavily on patient anatomy, surgeon preference, and the specific platform and instrumentation, these simulated layouts should be regarded as illustrative rather than prescriptive or universally applicable.

With this in mind, we summarize patient positioning and port layout for the three representative procedures below, focusing on patient posture, the number of camera/tool ports, and the overall port arrangement (Fig.~\ref{fig:port}). Since the skin model does not capture abdominal insufflation, we apply a $\sim$4\,cm outward offset---so ports may appear slightly above the skin surface---while preserving the intended distances between the ports and the target region.

\textbf{Robotic cholecystectomy:} Patient is typically placed in reverse Trendelenburg ($\sim$15$^\circ$--45$^\circ$) to move the intestines away from the surgical working space in the right upper abdomen. 1 camera port + 2 surgical tool ports and 1 retraction port are typically used. The camera port is placed at/near the umbilicus, with tool and retraction ports placed $\sim$6 - 8\,cm to each other. Ports are typically arranged in either a line or arc centered on the approximate gallbladder position. To allow for the additional space for dissection and gallbladder mobilization, the workspace was expanded $\sim$2\,cm beyond the gallbladder surface~\cite{changRobotic2020, marescauxTelerobotic2001}.

\textbf{Robotic inguinal hernia repair:} Patient is placed at ($\sim$15$^\circ$--30$^\circ$) Trendelenburg to move the intestines away from the surgical working space in the lower abdomen. 1 camera port + 2 surgical tool ports are typically used. The camera port is placed near the umbilicus. Two tool ports are placed 6-8 cm from the camera port to achieve triangulation toward the involved groin. The surgical workspace is a ~10 × 10 cm posterior inguinal ROI centered on the myopectineal orifice and deep inguinal ring, extending ~4–5 cm cranially to include the peritoneal flap.~\cite{waiteComparison2016, morrellRobotic2021}.

\textbf{Robotic sleeve gastrectomy}: Patient is placed in reverse Trendelenburg ($\sim$15$^\circ$--45$^\circ$) to move the transverse colon and small intestines away from the surgical working space in the left upper abdomen. 1 camera port + 3 surgical tool ports are used. An additional laparoscopic assistant port may also be used in cases. Camera port is typically placed near the umbilicus, additional ports are placed $\sim$ 6-8 cm apart and with two left abdominal ports placed for access to the left upper abdomen. The workspace is selected to cover the $\sim$22×15 cm stomach footprint, plus an added region reaching to the hiatus for allow for identification of the left crus and esophagus during the procedure~\cite{diasROBOTIC2020, kostovComparison2022}.

\begin{figure*}
    \centering
    \includegraphics[width=0.95\linewidth]{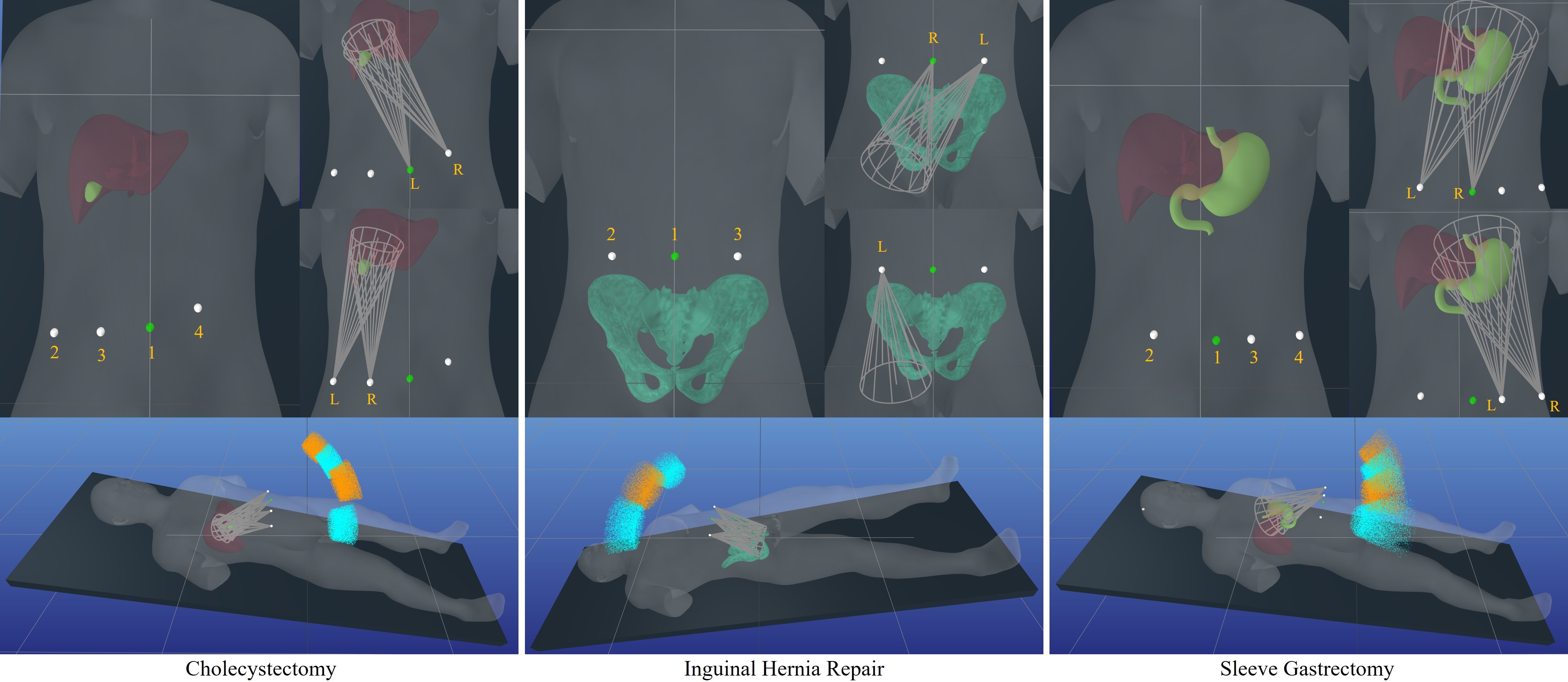}
    \caption{Port placement layouts and workspace of three representative robotic laparoscopic procedures. Green/white points: endoscope/wristed-tool ports with port numbers and assigned to robot L/R arm; gray cones: tool-tip task spaces. Cyan/orange point clouds: left/right robot end-effector target poses.}
    \label{fig:port}
\end{figure*}

}

\subsection{Humanoid Platforms}
Table~\ref{tab:cmp-humanoids} summarizes the height, arm DoF, arm length, and payload of the three platforms. Although all are Unitree robots, they span short to tall with notable differences in arm span and kinematic redundancy, which shape the end-effector workspace and hence the reachability metrics reported here; our comparison thus generalizes to platforms with similar geometry and kinematics.

\begin{figure}[t]
    \centering
    \includegraphics[width=0.9\linewidth]{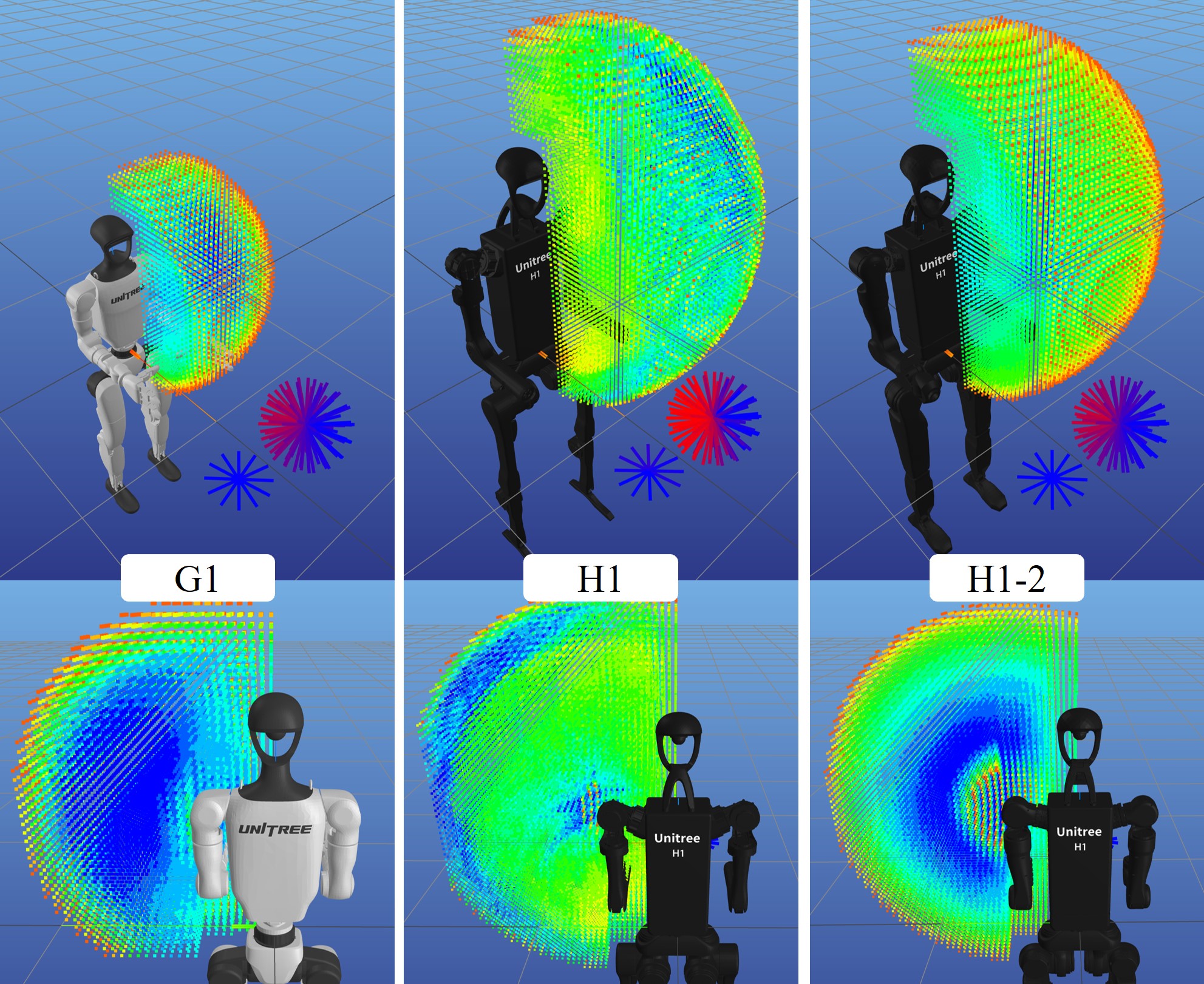}
    \caption{Capability maps of three humanoid platforms. Blue/Red: high/low numbers of joint configuration reaching that voxel or direction.}
    \label{fig:capability_maps}
\end{figure}

Fig.~\ref{fig:capability_maps} shows the arm end-effector capability maps of the three humanoids. We discretize the workspace into 2\,cm voxels with 100 approach directions and 12 roll bins. Each voxel is colored by the number of feasible joint configurations that reach it, reflecting kinematic redundancy; we also aggregate counts over approach and roll to visualize directional reachability. {H1} and {H1-2} span a larger workspace than {G1}, but {H1} attains lower reachability due to fewer arm DoFs. All platforms show higher reachability in the body-front approaching direction; {G1} and {H1-2} have more uniform roll reachability than H1.

\begin{table}
\centering
\caption{Specifications of Evaluated Humanoid Platforms}
\label{tab:cmp-humanoids}
\begin{tblr}{
  width = \linewidth,
  colspec = {Q[200]Q[250]Q[200]Q[200]},
  cells = {c},
  hline{1-2,6} = {-}{},
}
Specs            & Unitree G1 (Edu) & Unitree H1 & Unitree H1-2 \\
Height       & 1.32~m             & 1.52~m       & 1.503~m         \\
Arm DoF          & 7                & 4          & 7             \\
Arm Length   & 0.45~m            & 0.338~m        & 0.685~m         \\
Arm Payload & 3~kg                & --         & 7~kg             
\end{tblr}
\end{table}

\subsection{Surgery Reachability Results Across Humanoid Platforms}

We model the tool-tip task space at each port as a cone (spherical sector) with apex at the RCM, parameterized by span $\theta$ and depth limits $r_{\min},r_{\max}$. With half-angle $\alpha=\theta/2$, its volume is
$V_{\mathrm{cone}}=\frac{\pi\tan^2\alpha}{3}\left(r_{\max}^3-r_{\min}^3\right)$.
To estimate reach rate, we sample $N=10^4$ tool-tip poses uniformly in the cone, map each to a tool-base target via~\eqref{eq:G_T0_star} and then to an end-effector target using the optimized mount~\eqref{eq:opt-result}, and mark a sample reachable if IK yields at least one collision-free solution.
We evaluate two kinds of tools: (i) an endoscope and (ii) active 3-DoF (pitch--yaw--roll) grippers ignoring gripper actuation DoF.
The tool base is rigidly mounted to the humanoid end-effector, and we discretize the mount pitch and yaw to $[-30^\circ,30^\circ]$ in $15^\circ$ steps (Fig.~\ref{fig:frame-mounting}).
Since the investigated procedures typically use $3$--$4$ ports, we deploy 2 robots per procedure, assigning each to a pair of neighboring ports or a single port to maintain feasible inter-robot spacing, as shown in Fig.~\ref{fig:port}.




To demonstrate the effectiveness of our optimization framework, we compare three methods: 
\textbf{1) Optimize Base\&Mount} with simultaneous optimization of robot base placement and tool mounting as proposed in Section~\ref{subsec:opt-framework}.
\textbf{2) Optimize Base} as a variant of (i) that optimizes only the robot base placement while keeping the default tool mounting.
\textbf{3) Heuristic baseline} that uses the default tool mounting and selects the robot base placement by aligning robot arm workspace center with mean tool-base position and facing the mean target direction. Given tool-base poses $\{{}^{\mathrm{W}}\!\mathbf{p}^{\mathrm{L}}_{i}, {}^{\mathrm{W}}\!\mathbf{R}^{\mathrm{L}}_{i}\}_{i=1}^{N_{\mathrm{L}}}$ and  $\{{}^{\mathrm{W}}\!\mathbf{p}^{\mathrm{R}}_{i}, {}^{\mathrm{W}}\!\mathbf{R}^{\mathrm{R}}_{i}\}_{i=1}^{N_{\mathrm{R}}}$, and left-arm reachability-map center in the base frame ${}^{\mathrm{B}}\mathbf{c}_{\mathrm{L}}=[c_x,c_y,c_z]^\top$, we compute targets midpoint ${}^{\mathrm{W}}\!\mathbf{p}_m=\tfrac{1}{2}(\frac{1}{N_{\mathrm{L}}}\sum_i {}^{\mathrm{W}}\!\mathbf{p}^{\mathrm{L}}_{i} + \frac{1}{N_{\mathrm{R}}}\sum_j {}^{\mathrm{W}}\!\mathbf{p}^{\mathrm{R}}_{j} )$, and mean tool $x$-axis direction $\bar{\mathbf{a}}=\frac{1}{N_{\mathrm{L}}+N_{\mathrm{R}}}\!\left(\sum_i {}^{\mathrm{W}}\!\mathbf{R}^{\mathrm{L}}_{i}\mathbf{e}_x+\sum_j {}^{\mathrm{W}}\!\mathbf{R}^{\mathrm{R}}_{j}\mathbf{e}_x\right)$.
Then the robot base translation ${}^{\mathrm{W}}\!\mathbf{p}_{\mathrm{B}}$ and yaw $\psi$ are given by
\begin{equation}
\begin{aligned}
     {}^{\mathrm{W}}\!\mathbf{p}_{\mathrm{B}} &=\big(\mathbf{p}_m-c_z\mathbf{e}_z\big)-c_x\frac{[\bar{a}_x,\bar{a}_y,0]^\top}{\sqrt{\bar{a}_x^2+\bar{a}_y^2}}, \\
        \psi  &=\mathrm{atan2}(\bar{a}_y,\bar{a}_x).
\end{aligned}   
\end{equation}

\begin{table*}
\centering
\caption{Comparison of Surgery Workspace Reachability }
\label{tab:cmp}
\begin{tblr}{
  width = \linewidth,
  colspec = {Q[110]Q[280]Q[1]Q[55]Q[55]Q[55]Q[55]Q[1]Q[55]Q[55]Q[55]Q[1]Q[55]Q[55]Q[55]Q[55]},
  cells = {c},
  cell{1}{1}  = {r=3}{},
  cell{1}{4}  = {c=4}{},
  cell{1}{9}  = {c=3}{},
  cell{1}{13} = {c=4}{},
  cell{4}{1}  = {r=3}{},
  cell{7}{1}  = {r=3}{},
  cell{10}{1} = {r=3}{},
  hline{1,4,7,10,13} = {-}{},
  hline{2} = {4-7,9-11,13-16}{},
}
Humanoid Platform & Surgery Type              &  & Cholecystectomy &        &        &        &  & Inguinal Hernia Repair &      &      &  & Gastrectomy &      &      &      \\
                  & Port \#                   &  & 1               & 2      & 3      & 4      &  & 1                      & 2    & 3    &  & 1           & 2    & 3    & 4    \\
                  & Workspace Volume (cm$^3$) &  & 532             & 532    & 392    & 532    &  & 865                    & 865  & 874  &  & 2290        & 3009 & 2290 & 2495 \\

Unitree G1 (Edu)  & Heuristic                 &  & 84.3\%          & 80.9\% & 70.2\% & 79.0\% &  & 60.5\%                 & 61.1\% & 78.1\% &  & 47.3\%      & 46.3\% & 52.3\% & 36.0\% \\
                  & Optimize Base             &  & 88.1\%          & 80.4\% & 82.0\% & 85.5\% &  & 78.9\%                 & 76.9\% & 88.3\% &  & 60.4\%      & 58.2\% & 62.6\% & 68.8\% \\
                  & Optimize Base+Mount        &  & 91.2\%          & 87.7\% & \textbf{90.7}\% & 87.6\% &                 & \textbf{87.3}\% & \textbf{89.4}\% & \textbf{94.2}\% &  & 69.6\%      & 71.7\% & \textbf{77.9}\% & 67.9\% \\

Unitree H1        & Heuristic                 &  & 10.6\%          & 1.6\%  & 3.5\%  & 0.5\%  &  & --                     & --    & 2.1\%  &  & 0.3\%       & 2.5\%  & 1.7\%  & 1.3\%  \\
                  & Optimize Base             &  & 4.5\%           & 3.0\%  & 8.2\%  & 1.7\%  &  & 0.9\%                  & 3.3\% & 7.8\%  &  & 1.2\%       & 1.5\%  & 1.6\%  & 3.1\%  \\
                  & Optimize Base+Mount        &  & 5.7\%           & 2.9\%  & 4.7\%  & 1.5\%  &  & 2.4\%                  & 2.9\% & 10.4\% &  & 1.2\%       & 2.0\%  & 2.1\%  & 3.5\%  \\

Unitree H1-2      & Heuristic                 &  & 70.2\%          & 49.3\% & 76.5\% & 49.2\% &  & 41.98\%                & 60.1\% & 80.9\% &  & 64.1\%      & 47.3\% & 41.7\% & 56.1\% \\
                  & Optimize Base             &  & 81.6\%          & 83.8\% & 88.6\% & 80.4\% &  & 79.4\%                 & 82.4\% & 82.7\% &  & 75.3\%      & 52.3\% & 49.2\% & 74.0\% \\
                  & Optimize Base+Mount        &  & \textbf{97.1}\%          & \textbf{91.9}\% & 90.4\% & \textbf{91.0}\% &  & 83.9\%                 & 82.9\% & 86.7\% &  & \textbf{85.9}\%      & \textbf{75.5}\% & 70.7\% & \textbf{86.6}\% \\
\end{tblr}
\end{table*}

\begin{figure*}
    \centering
    \includegraphics[width=0.99\linewidth]{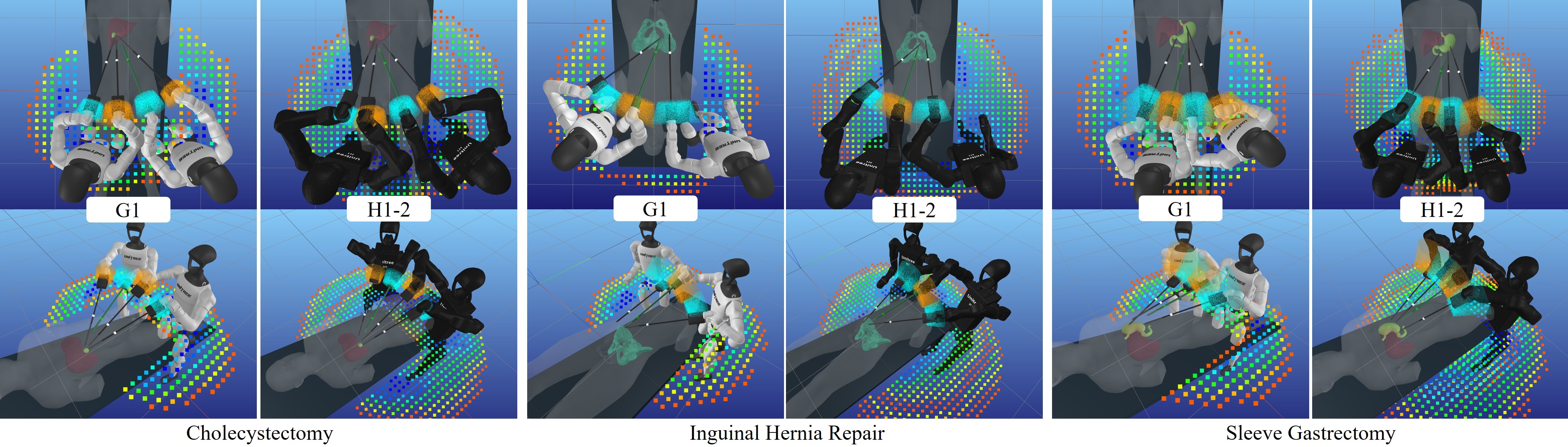}
    \caption{Robot base placement results of humanoids G1 and H1-2 for three representative robotic laparoscopic procedures. Left/Right end-effector targets: Cyan/Orange. Endoscope: Green; Wristed tool: Black. Floor tiles indicate base candidates colored by reachable-target count (blue high, red low).}
    \label{fig:placement}
\end{figure*}

Table~\ref{tab:cmp} reports the workspace reachability achieved by the three compared methods on three humanoid platforms, while Fig.~\ref{fig:placement} provides a qualitative overview of the corresponding optimized base placements. In Table~\ref{tab:cmp}, blank entries indicate that no feasible configuration was found because at least one environment constraint was violated (e.g., clearance/collision or placement bounds). 

Across all three procedures, the H1 platform consistently underperforms. We attribute the results primarily to its lower arm dexterity (4-DoF), which makes it difficult to satisfy the required end-effector orientations while reaching the sampled target set. Consequently, the discussion below focuses on G1 and H1-2 to provide a more informative comparison.

For robotic cholecystectomy, the required workspace volume is the smallest among the three procedures, and the desired end-effector workspaces are largely separated with minimal overlap. In this setting, the simultaneous optimization method achieves the highest reachability across all ports in Table~\ref{tab:cmp}. Moreover, H1-2 performs slightly better than G1, reaching over 90\% across ports.

For robotic inguinal hernia repair, the required workspace volume increases compared to cholecystectomy, but the port-specific workspaces remain mostly non-overlapping. The simultaneous optimization method again yields the best reachability across ports, while G1 and H1-2 exhibit comparable overall performance. A notable exception is Port~3, which shows higher reachability than other ports; this is expected because the optimization associated with this port effectively targets a single port, avoiding the trade-off required when balancing reachability across two ports simultaneously.

Sleeve gastrectomy imposes the most demanding workspace requirement: it has the largest workspace volume among the three procedures and contains substantial overlap between port workspaces. Although the simultaneous optimization method remains the best-performing strategy across ports, the absolute reachability is markedly lower than in the other two procedures, reflecting the increased geometric constraints. In addition, Fig.~\ref{fig:placement} shows that the optimized humanoid bases become spatially crowded for this procedure. Since our reachability metric does not yet account for inter-arm interference or collision between multiple humanoids, the reported reachability should be interpreted as an optimistic upper bound; incorporating these interactions may further reduce the effective reachability in practice.
Moreover, placing two bedside robots leaves no room for a human assistant to operate the fifth port in sleeve gastrectomy, highlighting a practical limitation of our optimized setup in this case.

{
\subsection{Discussion: Humanoid Potential for Surgical Tasks}
\label{subsec:discussion}


\textbf{Can a humanoid perform a surgical task?}
Our results suggest humanoids are most promising for procedures with smaller task spaces and minimal overlap between tool workspaces (e.g., cholecystectomy), where we observe around 90\% reachability across ports. In contrast, for procedures requiring larger task spaces with overlapping regions and many ports (e.g., sleeve gastrectomy), the evaluated humanoids do not yet show reliable coverage. Additionally, the limitation of each humanoid to bimanual operation poses a logistic challenge for base placement when surgeries require more than 4 ports. Moreover, high reachability should be interpreted as an optimistic upper bound: a more rigorous evaluation of dexterity control and manipulation such as pose tracking is needed. 
Broader topics such as control accuracy, multi-agent coordination and safety are equally important and lie beyond the scope of this kinematic study.
Finally, the near 95\% reachability in single-port optimization indicates a plausible near-term role for humanoids as a dedicated assistant operating through a single auxiliary port.

\textbf{What enables higher capability?}
The poor performance of H1 highlights the importance of arm DoF; limited dexterity restricts end-effector orientations even when positions are reachable. A slightly larger body/arm scale can improve coverage, but it also makes multi-robot deployment harder due to increased risk of inter-arm interference, collisions around the patient and insufficient space for a human assistant.

\textbf{What is still missing?}
Our port placements follow clinical practice for dedicated surgical systems (e.g., \textit{da Vinci} surgical system), which may not be optimal for humanoid kinematics. Humanoids likely require port layouts with appropriate inter-port distances and angular spread for bimanual operation, and multi-agent scenarios further require collision-aware placement and coordination. Moreover, conventional surgeon-ergonomic triangulation port layouts will be evaluated in future work.
Finally, we fix the humanoid base during the optimization; exploiting whole-body locomotion during a procedure is a promising direction that could further enlarge the reachable workspace and distinguish humanoids from fixed bimanual manipulators.

}
\section{CONCLUSIONS}
We present a capability-map-based optimization framework to study humanoids for robot-assisted surgery, with a focus on tool-tip reachability under RCM constraints and bimanual coordination. By decoupling robot reachability from tool kinematics, our framework is adaptable across tools and platforms. We developed simultaneous optimization over humanoid base placement and tool mounting orientations, which consistently outperformed a base-only variant and a heuristic baseline across the three representative laparoscopic procedures.
We evaluate the framework on three humanoids spanning a wide range of body dimensions and kinematic redundancy across three representative surgical setups. For procedures with smaller, largely non-overlapping port workspaces (e.g., cholecystectomy and inguinal hernia repair), humanoids can achieve high reachability (around 90\%), suggesting practical potential for selected procedures and for single-port assistant roles. In contrast, larger workspaces with substantial overlap and many ports (e.g., sleeve gastrectomy) remain difficult and yield lower reachability.

\bibliographystyle{IEEEtran}
\bibliography{citation}

@misc{atarHumanoids2025,
  title = {Humanoids in {{Hospitals}}: {{A Technical Study}} of {{Humanoid Robot Surrogates}} for {{Dexterous Medical Interventions}}},
  shorttitle = {Humanoids in {{Hospitals}}},
  author = {Atar, Soofiyan and Liang, Xiao and Joyce, Calvin and Richter, Florian and Ricardo, Wood and Goldberg, Charles and Suresh, Preetham and Yip, Michael},
  year = 2025,
  month = jul,
  number = {arXiv:2503.12725},
  eprint = {2503.12725},
  primaryclass = {cs},
  publisher = {arXiv},
  doi = {10.48550/arXiv.2503.12725},
  urldate = {2026-02-27},
  archiveprefix = {arXiv}
}

@article{bornerHuman2025,
  title = {Human {{BioMolecular Atlas Program}} ({{HuBMAP}}): {{3D Human Reference Atlas}} Construction and Usage},
  shorttitle = {Human {{BioMolecular Atlas Program}} ({{HuBMAP}})},
  author = {B{\"o}rner, Katy and Blood, Philip D. and Silverstein, Jonathan C. and Ruffalo, Matthew and Satija, Rahul and Teichmann, Sarah A. and Pryhuber, Gloria J. and Misra, Ravi S. and Purkerson, Jeffrey M. and Fan, Jean and Hickey, John W. and Molla, Gesmira and Xu, Chuan and Zhang, Yun and Weber, Griffin M. and Jain, Yashvardhan and Qaurooni, Danial and Kong, Yongxin and Bueckle, Andreas and Herr, Bruce W.},
  year = 2025,
  month = apr,
  journal = {Nature Methods},
  volume = {22},
  number = {4},
  pages = {845--860},
  publisher = {Nature Publishing Group},
  issn = {1548-7105},
  doi = {10.1038/s41592-024-02563-5},
  urldate = {2026-02-28},
  copyright = {2025 The Author(s)},
  langid = {english}
}

@inproceedings{burgetStance2015,
  title = {Stance Selection for Humanoid Grasping Tasks by Inverse Reachability Maps},
  booktitle = {2015 {{IEEE International Conference}} on {{Robotics}} and {{Automation}} ({{ICRA}})},
  author = {Burget, Felix and Bennewitz, Maren},
  year = 2015,
  month = may,
  pages = {5669--5674},
  issn = {1050-4729},
  doi = {10.1109/ICRA.2015.7139993},
  urldate = {2026-01-25}
}

@article{changRobotic2020,
  title = {Robotic {{Biliary Surgery}}},
  author = {Chang, Karen and Gokcal, Fahri and Kudsi, Omar Yusef},
  year = 2020,
  month = apr,
  journal = {Surgical Clinics of North America},
  series = {Robotic {{Surgery}}},
  volume = {100},
  number = {2},
  pages = {283--302},
  issn = {0039-6109},
  doi = {10.1016/j.suc.2019.12.002},
  urldate = {2026-02-23}
}

@article{diasROBOTIC2020,
  title = {{{ROBOTIC GASTRECTOMY}}: {{TECHNIQUE STANDARDIZATION}}},
  shorttitle = {{{ROBOTIC GASTRECTOMY}}},
  author = {Dias, Andre Roncon and Ramos, Marcus Fernando Kodama Pertille and Szor, Daniel Jose and Abdalla, Ricardo and Barchi, Leandro and Yagi, Osmar Kenji and {Ribeiro-Junior}, Ulysses and Zilberstein, Bruno and Cecconello, Ivan},
  year = 2020,
  journal = {ABCD. Arquivos Brasileiros de Cirurgia Digestiva (S\~ao Paulo)},
  volume = {33},
  pages = {e1542},
  publisher = {Col\'egio Brasileiro de Cirurgia Digestiva},
  issn = {0102-6720, 2317-6326},
  doi = {10.1590/0102-672020200003e1542},
  urldate = {2026-02-28},
  langid = {english}
}

@inproceedings{dongOrientationbased2015,
  title = {Orientation-Based Reachability Map for Robot Base Placement},
  booktitle = {2015 {{IEEE}}/{{RSJ International Conference}} on {{Intelligent Robots}} and {{Systems}} ({{IROS}})},
  author = {Dong, Jun and Trinkle, Jeffrey C.},
  year = 2015,
  month = sep,
  pages = {1488--1493},
  doi = {10.1109/IROS.2015.7353564},
  urldate = {2026-01-29}
}

@misc{hanARport2026,
  title = {{{ARport}}: {{An Augmented Reality System}} for {{Markerless Image-Guided Port Placement}} in {{Robotic Surgery}}},
  shorttitle = {{{ARport}}},
  author = {Han, Zheng and Yang, Zixin and Long, Yonghao and Zhang, Lin and Kazanzides, Peter and Dou, Qi},
  year = 2026,
  month = feb,
  number = {arXiv:2602.14153},
  eprint = {2602.14153},
  primaryclass = {cs},
  publisher = {arXiv},
  doi = {10.48550/arXiv.2602.14153},
  urldate = {2026-02-27},
  archiveprefix = {arXiv}
}

@article{jauhriRobot2022,
  title = {Robot {{Learning}} of {{Mobile Manipulation With Reachability Behavior Priors}}},
  author = {Jauhri, Snehal and Peters, Jan and Chalvatzaki, Georgia},
  year = 2022,
  month = jul,
  journal = {IEEE Robotics and Automation Letters},
  volume = {7},
  number = {3},
  pages = {8399--8406},
  issn = {2377-3766},
  doi = {10.1109/LRA.2022.3188109},
  urldate = {2026-01-25},
  langid = {american}
}

@article{kostovComparison2022,
  title = {Comparison of Short Term Results Following Robotic and Laparoscopic Total Gastrectomy and {{D2}} Lymph Node Dissection},
  author = {Kostov, Gancho and Dimov, Rossen and Doykov, Mladen},
  year = 2022,
  month = dec,
  journal = {Folia Medica},
  volume = {64},
  number = {6},
  pages = {889--895},
  publisher = {Plovdiv Medical University},
  issn = {1314-2143, 0204-8043},
  doi = {10.3897/folmed.64.e89545},
  urldate = {2026-02-28},
  copyright = {2022 Gancho Kostov, Rossen Dimov, Mladen Doykov},
  langid = {english}
}

@article{kyrariniSurvey2021,
  title = {A {{Survey}} of {{Robots}} in {{Healthcare}}},
  author = {Kyrarini, Maria and Lygerakis, Fotios and Rajavenkatanarayanan, Akilesh and Sevastopoulos, Christos and Nambiappan, Harish Ram and Chaitanya, Kodur Krishna and Babu, Ashwin Ramesh and Mathew, Joanne and Makedon, Fillia},
  year = 2021,
  month = mar,
  journal = {Technologies},
  volume = {9},
  number = {1},
  pages = {8},
  publisher = {Multidisciplinary Digital Publishing Institute},
  issn = {2227-7080},
  doi = {10.3390/technologies9010008},
  urldate = {2026-02-27},
  copyright = {http://creativecommons.org/licenses/by/3.0/},
  langid = {english}
}

@misc{liangAutonomous2025,
  title = {Towards {{Autonomous Tape Handling}} for {{Robotic Wound Redressing}}},
  author = {Liang, Xiao and Shen, Lu and Zhang, Peihan and Atar, Soofiyan and Richter, Florian and Yip, Michael},
  year = 2025,
  month = oct,
  number = {arXiv:2510.06127},
  eprint = {2510.06127},
  primaryclass = {cs},
  publisher = {arXiv},
  doi = {10.48550/arXiv.2510.06127},
  urldate = {2026-02-27},
  archiveprefix = {arXiv}
}

@misc{liangLapSurgie2026,
  title = {{{LapSurgie}}: {{Humanoid Robots Performing Surgery}} via {{Teleoperated Handheld Laparoscopy}}},
  shorttitle = {{{LapSurgie}}},
  author = {Liang, Zekai and Liang, Xiao and Atar, Soofiyan and Das, Sreyan and Chiu, Zoe and Zhang, Peihan and Joyce, Calvin and Richter, Florian and Liu, Shanglei and Yip, Michael C.},
  year = 2026,
  month = feb,
  number = {arXiv:2510.03529},
  eprint = {2510.03529},
  primaryclass = {cs},
  publisher = {arXiv},
  doi = {10.48550/arXiv.2510.03529},
  urldate = {2026-02-27},
  archiveprefix = {arXiv}
}

@article{marescauxTelerobotic2001,
  title = {Telerobotic {{Laparoscopic Cholecystectomy}}: {{Initial Clinical Experience With}} 25 {{Patients}}},
  shorttitle = {Telerobotic {{Laparoscopic Cholecystectomy}}},
  author = {Marescaux, Jacques and Smith, Michelle K. and F{\"o}lscher, Daniel and Jamali, Faek and Malassagne, Benoit and Leroy, Joel},
  year = 2001,
  month = jul,
  journal = {Annals of Surgery},
  volume = {234},
  number = {1},
  pages = {1--7},
  issn = {0003-4932},
  doi = {10.1097/00000658-200107000-00001},
  urldate = {2026-02-28},
  pmcid = {PMC1421940},
  pmid = {11420476}
}

@article{morrellRobotic2021,
  title = {Robotic {{TAPP}} Inguinal Hernia Repair: Lessons Learned from 97 Cases},
  shorttitle = {Robotic {{TAPP}} Inguinal Hernia Repair},
  author = {Morrell, Andre Luiz Gioia and Morrell Junior, Alexander Charles and Mendes, Jose Mauricio Freitas and Morrell, Allan Gioia and Morrell, Alexander},
  year = 2021,
  journal = {Revista do Col\'egio Brasileiro de Cirurgi\~oes},
  volume = {48},
  pages = {e20202704},
  publisher = {Col\'egio Brasileiro de Cirurgi\~oes},
  issn = {0100-6991, 1809-4546},
  doi = {10.1590/0100-6991e-20202704},
  urldate = {2026-02-28},
  langid = {english}
}

@article{osburgIncreasing2025,
  title = {Increasing {{Reachability}} in {{Robotic Ultrasound Through Base Placement}} and {{Tool Design}}},
  author = {Osburg, Jonas and Nguyen, Ngoc Thinh and Ernst, Floris},
  year = 2025,
  journal = {The International Journal of Medical Robotics and Computer Assisted Surgery},
  volume = {21},
  number = {1},
  pages = {e70037},
  issn = {1478-596X},
  doi = {10.1002/rcs.70037},
  urldate = {2026-01-27},
  langid = {english}
}

@inproceedings{porgesReachability2014,
  title = {Reachability and {{Capability Analysis}} for {{Manipulation Tasks}}},
  booktitle = {{{ROBOT2013}}: {{First Iberian Robotics Conference}}},
  author = {Porges, Oliver and Stouraitis, Theodoros and Borst, Christoph and Roa, Maximo A.},
  editor = {Armada, Manuel A. and Sanfeliu, Alberto and Ferre, Manuel},
  year = 2014,
  pages = {703--718},
  publisher = {Springer International Publishing},
  address = {Cham},
  doi = {10.1007/978-3-319-03653-3_50},
  isbn = {978-3-319-03653-3},
  langid = {english}
}

@article{silvera-tawilRobotics2024,
  title = {Robotics in {{Healthcare}}: {{A Survey}}},
  shorttitle = {Robotics in {{Healthcare}}},
  author = {{Silvera-Tawil}, David},
  year = 2024,
  month = jan,
  journal = {SN Computer Science},
  volume = {5},
  number = {1},
  pages = {189},
  issn = {2661-8907},
  doi = {10.1007/s42979-023-02551-0},
  urldate = {2026-02-27},
  langid = {english}
}

@article{sundaramTaskspecific2022,
  title = {Task-Specific Robot Base Pose Optimization for Robot-Assisted Surgeries},
  author = {Sundaram, Ashok M. and Budjakoski, Nikola and Klodmann, Julian and Roa, M{\'a}ximo A.},
  year = 2022,
  month = dec,
  journal = {Frontiers in Robotics and AI},
  volume = {9},
  pages = {899646},
  issn = {2296-9144},
  doi = {10.3389/frobt.2022.899646},
  urldate = {2026-01-26},
  langid = {american},
  pmcid = {PMC9755868},
  pmid = {36530494}
}

@article{trabelsiRobot2024,
  title = {Robot Base Placement and Tool Mounting Optimization Based on Capability Map for Robot-Assistant Camera Holder},
  author = {Trabelsi, Amir and Sandoval, Juan and Mlika, Abdelfattah and Lahouar, Samir and Zeghloul, Said and Laribi, Med Amine},
  year = 2024,
  month = aug,
  journal = {Robotica},
  volume = {42},
  number = {8},
  pages = {2489--2510},
  issn = {0263-5747, 1469-8668},
  doi = {10.1017/S0263574724000870},
  urldate = {2026-01-27},
  langid = {english}
}

@inproceedings{vahrenkampRobot2013,
  title = {Robot Placement Based on Reachability Inversion},
  booktitle = {2013 {{IEEE International Conference}} on {{Robotics}} and {{Automation}}},
  author = {Vahrenkamp, Nikolaus and Asfour, Tamim and Dillmann, R{\"u}diger},
  year = 2013,
  month = may,
  pages = {1970--1975},
  issn = {1050-4729},
  doi = {10.1109/ICRA.2013.6630839},
  urldate = {2026-01-27}
}

@article{waiteComparison2016,
  title = {Comparison of Robotic versus Laparoscopic Transabdominal Preperitoneal ({{TAPP}}) Inguinal Hernia Repair},
  author = {Waite, Kimberly E. and Herman, Mark A. and Doyle, Patrick J.},
  year = 2016,
  month = sep,
  journal = {Journal of Robotic Surgery},
  volume = {10},
  number = {3},
  pages = {239--244},
  issn = {1863-2491},
  doi = {10.1007/s11701-016-0580-1},
  urldate = {2026-02-22},
  langid = {english}
}

@inproceedings{xuOptimal2021,
  title = {Optimal {{Multi-Manipulator Arm Placement}} for {{Maximal Dexterity}} during {{Robotics Surgery}}},
  booktitle = {2021 {{IEEE International Conference}} on {{Robotics}} and {{Automation}} ({{ICRA}})},
  author = {Xu, Mingwei and Di, James and Das, Nikhil and Yip, Michael C.},
  year = 2021,
  month = may,
  pages = {9752--9758},
  issn = {2577-087X},
  doi = {10.1109/ICRA48506.2021.9561570},
  urldate = {2026-01-25}
}

@inproceedings{yoonOptimizing2024,
  title = {Optimizing {{Base Placement}} of {{Surgical Robot}}: {{Kinematics Data-Driven Approach}} by {{Analyzing Working Pattern}}},
  shorttitle = {Optimizing {{Base Placement}} of {{Surgical Robot}}},
  booktitle = {2024 {{IEEE}}/{{RSJ International Conference}} on {{Intelligent Robots}} and {{Systems}} ({{IROS}})},
  author = {Yoon, Jeonghyeon and Park, Junhyun and Park, Hyojae and Lee, Hakyoon and Lee, Sangwon and Hwang, Minho},
  year = 2024,
  month = oct,
  pages = {6907--6914},
  issn = {2153-0866},
  doi = {10.1109/IROS58592.2024.10802398},
  urldate = {2026-01-27}
}

@article{zachariasCapability2013,
  title = {The Capability Map: A Tool to Analyze Robot Arm Workspaces},
  shorttitle = {The Capability Map},
  author = {Zacharias, Franziska and Borst, Christoph and Wolf, Sebastian and Hirzinger, Gerd},
  year = 2013,
  month = dec,
  journal = {International Journal of Humanoid Robotics},
  volume = {10},
  number = {04},
  pages = {1350031},
  publisher = {World Scientific Publishing Co.},
  issn = {0219-8436},
  doi = {10.1142/S021984361350031X},
  urldate = {2026-01-26}
}

@inproceedings{zachariasCapturing2007,
  title = {Capturing Robot Workspace Structure: Representing Robot Capabilities},
  shorttitle = {Capturing Robot Workspace Structure},
  booktitle = {2007 {{IEEE}}/{{RSJ International Conference}} on {{Intelligent Robots}} and {{Systems}}},
  author = {Zacharias, Franziska and Borst, Christoph and Hirzinger, Gerd},
  year = 2007,
  month = oct,
  pages = {3229--3236},
  issn = {2153-0866},
  doi = {10.1109/IROS.2007.4399105},
  urldate = {2026-01-26}
}

@inproceedings{zhangConfiguration2021,
  title = {Configuration, {{Layout}}, and {{Pose Optimization}} of {{Surgical Robotic System}}},
  booktitle = {2021 27th {{International Conference}} on {{Mechatronics}} and {{Machine Vision}} in {{Practice}} ({{M2VIP}})},
  author = {Zhang, Xue and Xian, Yitian and Li, Jian and Yan Chiu, Philip Wai and Li, Zheng},
  year = 2021,
  month = nov,
  pages = {766--770},
  doi = {10.1109/M2VIP49856.2021.9665151},
  urldate = {2026-01-27}
}

\addtolength{\textheight}{-0cm}   






\end{document}